\documentclass[11pt,a4paper]{article} 
\usepackage[hyperref]{naaclhlt2019}
\usepackage{times}
\usepackage{latexsym}
\usepackage{tikz}
\usepackage{dingbat}
\usepackage{caption}
\usepackage{wasysym}
\usepackage{comment}
\usepackage[normalem]{ulem}
\usepackage{booktabs}

\aclfinalcopy 

\title{Probing What Different NLP Tasks Teach Machines\\ about Function Word Comprehension}

\author{Najoung Kim$^{\dagger,*}$, Roma Patel$^\phi$, Adam Poliak$^\dagger$, Alex Wang$^\partial$,\\\textbf{Patrick Xia$^\dagger$, R. Thomas McCoy$^\dagger$, Ian Tenney$^\Delta$, Alexis Ross$^\diamond$,}\\\textbf{Tal Linzen$^\dagger$, Benjamin Van Durme$^{\dagger}$, Samuel R. Bowman$^\partial$, Ellie Pavlick$^{\phi,}$\footnotemark}\\
  $^\dagger$Johns Hopkins University  $^\phi$Brown University\\
  $^\partial$New York University $^\Delta$Google AI Language $^\diamond$Harvard University\\
  }

\date{}
\begin{document}
\maketitle
\begin{abstract}

We introduce a set of nine challenge tasks that test for the understanding of function words. These tasks are created by structurally mutating sentences from existing datasets to target the comprehension of specific types of function words (e.g., prepositions, \textit{wh}-words). Using these probing tasks, we explore the effects of various pretraining objectives for sentence encoders (e.g., language modeling, CCG supertagging and natural language inference (NLI)) on the learned representations. Our results show that pretraining on language modeling performs the best on average across our probing tasks, supporting its widespread use for pretraining state-of-the-art NLP models, and CCG supertagging and NLI pretraining perform comparably. Overall, no pretraining objective dominates across the board, and our function word probing tasks highlight several intuitive differences between pretraining objectives, e.g., that NLI helps the comprehension of negation.

\end{abstract}

\section{Introduction}

Many recent advances in NLP have been driven by new approaches to representation learning---i.e., the design of models whose primary aim is to yield representations of words or sentences that are useful for a range of downstream applications \cite{RepEval:2017}. Approaches to representation learning typically differ in the architecture of the model used to learn the representations, the objective used to train that network, or both. Varying these factors can significantly impact performance on a broad range of NLP tasks \cite{mccann2017learned,peters2018deep,devlin2018bert}.

\begin{table*}[ht!]
\centering
\small
\begin{tabular}{lllc}
\toprule
    \multicolumn{4}{c}{Acceptability} \\\midrule
    \textit{wh}-word 	&                                 
        \multicolumn{2}{c}{\colorbox{GreenYellow}{\textbf{why}} are you so chippy about posh people?} & \textbf{\checkmark}\\
            & \multicolumn{2}{c}{$\dots$a Mr. Nice Guy like Melcher, \colorbox{GreenYellow}{\textbf{what}} is now 46} & \textbf{X}  \\		\midrule

    Def. 		&  \multicolumn{2}{c}{$\dots$ \colorbox{GreenYellow}{\textbf{the}} case is remarkable for \colorbox{GreenYellow}{\textbf{the}} cooperation $\dots$}   &   \textbf{\checkmark}	\\
              & \multicolumn{2}{c}{$\dots$\colorbox{GreenYellow}{\textbf{a}} case is remarkable for \colorbox{GreenYellow}{\textbf{a}} cooperation $\dots$} & \textbf{X}  \\	\midrule
    Coord. 	 	&  \multicolumn{2}{c}{I have also tried monthly data \colorbox{GreenYellow}{\textbf{and}} the results are the same.} &  \textbf{\checkmark}	\\
              &  \multicolumn{2}{c}{Rooms very clean \colorbox{GreenYellow}{\textbf{but}} smelled very fresh.} &  \textbf{X}    \\	\midrule
EOS 		&  \multicolumn{2}{c}{the forehead is gathered in a frown \colorbox{GreenYellow}{\textbf{//}} the mouth is slightly parted to reveal the teeth} &  \textbf{\checkmark} \\
             & \multicolumn{2}{c}{the forehead is gathered in a frown the mouth \colorbox{GreenYellow}{\textbf{//}} is slightly parted to reveal the teeth} & \textbf{X} 
				\\ 
				\toprule
		\multicolumn{4}{c}{NLI} \\\midrule  
Prep. & 	\colorbox{GreenYellow}{\textbf{With}} a single jerk the man's head tore free. & $\to$ \hspace{0.2cm} The man's head tore free  \colorbox{GreenYellow}{\textbf{from}} a single jerk. & \textbf{\checkmark} \\
    &  \colorbox{GreenYellow}{\textbf{With}} a single jerk the man's head tore free. & $\to$ \hspace{0.2cm} The man's head tore free \colorbox{GreenYellow}{\textbf{without}} a single jerk. & \textbf{X}\\\midrule

Negation	&This is a common problem. & $\to$ \hspace{0.2cm} This is            \colorbox{GreenYellow}{\textbf{not}} an \colorbox{GreenYellow}{\textbf{uncommon}} issue we are facing. &	\textbf{\checkmark}  \\
		& This is \colorbox{GreenYellow}{\textbf{not}} a common problem. & $\to$ \hspace{0.2cm} This is \colorbox{GreenYellow}{\textbf{not}} an \colorbox{GreenYellow}{\textbf{uncommon}} issue we are facing. & 	\textbf{X} \\\midrule
Spatial &	  To reach  $\dots$ turn \colorbox{GreenYellow}{\textbf{left}} up a small alleyway & $\to$ \hspace{0.2cm} do not turn \colorbox{GreenYellow}{\textbf{right}} up the alleyway $\dots$ & \textbf{\checkmark} \\ 
& To reach  $\dots$ turn \colorbox{GreenYellow}{\textbf{left}} up a small alleyway & $\to$ \hspace{0.2cm} Turn \colorbox{GreenYellow}{\textbf{right}} up the alleyway $\dots$ &\textbf{X}  \\ \midrule
Quant. &  \colorbox{GreenYellow}{\textbf{all}} taken up yeah & $\to$ \hspace{0.2cm} There are not still \colorbox{GreenYellow}{\textbf{some}} left  & \textbf{\checkmark}	\\
		&	\colorbox{GreenYellow}{\textbf{all}} taken up yeah& $\to$ \hspace{0.2cm}  There are still \colorbox{GreenYellow}{\textbf{some}} left & \textbf{X} 			\\\midrule
 
Comp.	&  Today there are \colorbox{GreenYellow}{\textbf{more}} than 300,000. & $\to$ \hspace{0.2cm}  Today there are not \colorbox{GreenYellow}{\textbf{less}} than 300,000. & \textbf{\checkmark}\\
			& Today there are \colorbox{GreenYellow}{\textbf{more}} than 300,000. & $\to$ \hspace{0.2cm} Today there are \colorbox{GreenYellow}{\textbf{less}} than 300,000. &\textbf{X} 	\\\bottomrule

\end{tabular}

\caption{Examples of sentences and sentence pairs corresponding to each of our probing datasets. The highlighted words are those that are relevant to the phenomena targeted by each set.}
\label{table:probe-examples}
\end{table*}

This paper investigates the role of pretraining objectives of sentence encoders, with respect to their capacity to understand function words (e.g., prepositions, conjunctions). Although the importance of finding an effective pretraining objective for learning better (or more generalizable) representations is well acknowledged, relatively few studies offer a controlled comparison of diverse pretraining objectives, holding model architecture constant.

We ask whether the linguistic properties implicitly captured by pretraining objectives measurably affect the types of linguistic information encoded in the learned representations. To this end, we explore whether qualitatively different objectives lead to demonstrably different sentence representations. We focus our analysis on function words because they play a key role in compositional meaning---e.g., introducing and identifying discourse referents or representing relationships between entities or ideas---and are not yet considered to be well-modeled by distributional semantics \cite{bernardi-EtAl:2015:LSDSem}. Our results suggest that different pretraining objectives give rise to differences in function word comprehension; for instance, we see that natural language inference helps understanding negation, and CCG supertagging helps recognizing meaningful sentence boundaries. However, overall, we find that the observed differences are not always straightforwardly interpretable, and further investigation is needed to determine which specific aspects of pretraining tasks yield good representations of function words. 

The analyses we present contribute new results in an ongoing line of research aimed at providing a finer-grained understanding of what neural networks capture about linguistic structure \citep[][\textit{i.a.}]{conneau2018you,poliak2018collecting,Blackbox:2018,tenney2019what}. Our contributions are: 

\begin{itemize}
\item We provide an in-depth exploration into how different pretraining objectives for sentence encoders affect the information encoded by the output representations. We isolate the effects of different pretraining objectives by holding the model architecture constant.
\item We study function words, which have been under-studied in previous works on representation learning, but are critical to language understanding.
\item We release nine new datasets,\footnote{The datasets are released as part of the Diverse Natural Language Inference Collection \citep[DNC,][]{poliak2018collecting}, available at \url{http://decomp.io}.} quality-controlled by both linguists and non-linguist annotators, to facilitate ongoing work and follow-up analysis. 
\end{itemize}

\section{Function Word Probing Tasks}
\label{sec:probing-tasks}

\subsection{Approach}

We introduce nine new probing tasks aimed at evaluating models' understanding of function words. We focus on function words because although they are key building blocks of compositional meaning and are highly frequent, they have received relatively little attention in the probing literature and in the distributional semantics literature. Each task targets the understanding of a specific type of function word; illustrative examples are given in Table~\ref{table:probe-examples}. Our expectation is that different pretraining objectives (see Section~\ref{sec:pretraining-tasks}) will yield sentence representations which measurably differ in their performance on these probing tasks.

We use two different formats for our probing tasks: acceptability judgment and natural language inference (NLI). The former uses a binary classification approach (acceptable/unacceptable) for probing a single sentence vector, in line with works such as \citet{conneau2018you} and \citet{adi2017fine}. The latter uses an entailment-based approach similar to \citet{white2017inference} and \citet{poliak2018collecting}, which is a ternary classification task (\textit{entailment}, \textit{contradiction}, \textit{neutral}) over sentence pairs. The format is selected based on the suitability to the particular function word type in question.

To generate our probing datasets, we make structural modifications to sentences drawn from existing corpora, targeting a particular type of function word. We heuristically apply modifications which we believe are likely to produce a specific label, and then recruit human annotators in order to produce the final labels used in our evaluations. The result is a publicly available suite of nine task datasets (four acceptability tasks and five NLI tasks) consisting of 3,544 annotated examples. Appendix~\ref{app:annotation} lists the sizes of each dataset. 

\subsection{Acceptability Judgment-Based Tasks}

We cast acceptability as a binary classification task following the format of such judgments commonly used in linguistics, in a similar manner to \citet{warstadt2018neural}. All tasks follow a common protocol of first identifying sentences that contain the construction that we are interested in, and then mutating half of the identified sentences to generate infelicitous versions of the original sentences. Unless stated otherwise, the original sentences are drawn from the test set of the Billion Word Benchmark \citep[BWB,][]{chelba2013one}.

\paragraph{Wh-Words} 
Understanding \textit{wh}-words (i.e., \textit{who, what, where, when, why, how}) depends on understanding the context and correctly identifying the antecedent, which may not be overtly present in the sentence. For instance, recognizing the infelicity of \textit{I talked about who I live} requires knowing that the (unstated) antecedent must be a place and not a person. Our dataset consists of sentences that contain one of the six \textit{wh}-words listed above. Half of these sentences are mutated versions of the original which are generated by replacing the original \textit{wh}-word with a different \textit{wh}-word randomly selected from the remaining five options. 

\paragraph{Definite-Indefinite Articles} The definiteness task probes the understanding of definiteness that arises by the use of the definite article (\textit{the}) versus indefinite articles (\textit{a} and \textit{an}). We find sentences containing multiple occurrences of \textit{the} or multiple occurrences of \textit{a}, and, for half of them, swap all such occurrences (i.e., replacing \textit{the} with \textit{a}\footnote{When we replace \textit{the} with \textit{a}, we choose \textit{a/an} as necessary based on the word it precedes.} or vice-versa). This gives us four types of sentences: unchanged sentences with multiple definite articles, unchanged sentences with multiple indefinite articles, sentences with all definite articles replaced by the indefinite article, and sentences with all indefinite articles replaced by the definite article. Our intent is that the former two types will be judged felicitous while the latter two will be infelicitous despite the fact that the sentence would be syntactically well-formed. We only focus on the cases with multiple occurrences of the same article, because replacing a single article most of the time did not significantly affect the acceptability (although it often did affect the actual meaning).

\paragraph{Coordinating Conjunctions} Correct understanding of coordinating conjunctions (\textit{and, but, or}) requires contextual comprehension of the two conjoined linguistic units, since different coordinating conjunctions express different logical relations, meaning their use is often restricted by the meanings of the conjoined items. We take sentences that contain coordinating conjunctions, and replace half of them with a version that contains a different conjunction. For example, the sentence \textit{Room's very clean \textbf{but} smelled very fresh} is infelicitous despite being syntactically well-formed; \textit{but} is unnatural here because the conjoined clauses do not form a clear contrast. Judging this sentence to be infelicitous requires a proper understanding of the ideas expressed in the clauses and how they relate to each other.

\paragraph{End-of-Sentence} The end-of-sentence (EOS) task tests a model's ability to identify semantically coherent chunks (i.e., sentences) in running text. In written text this is often indicated by punctuation marks such as periods, but humans are able to easily identify sentences even without overt markers. Thus, we take pairs of sentences from the same paragraph of the WikiText-103 \cite{merity2016pointer} test set and remove all punctuation marks and capitalization, and concatenate each sentence pair to create a line of running text.\footnote{We use WikiText instead of BWB because adjacent sentences in BWB are not logically contiguous and therefore may not be from the same discourse context. } Half of the dataset consists of a pair of valid sentences, and the other half consists of a pair of potentially invalid sentences generated from an incorrect segmentation of the running text, where the incorrect segmentation index is obtained by sampling from a Gaussian distribution centered around the correct index ($\sigma=2$) and rounding to the nearest integer.

\subsection{NLI-Based Tasks}
Our NLI-based probing tasks ask whether the choice of function word affects the inferences licensed by a sentence. 
These tasks consist of a pair of sentences---a premise $p$ and a hypothesis $h$---and ask whether or not $p$ entails $h$. We exploit the label changes induced by a targeted mutation of the sentence pairs taken from the Multi-genre Natural Language Inference dataset \citep[MNLI,][]{williams2018broad}. 
The rationale is that, if a change to a single function word in the premise changes the entailment label, that function word must play a significant role in the semantics of the sentence.

\paragraph{Prepositions} We manually curate a list of prepositions (see Appendix~\ref{app:preplist}) that are likely to be swapped with each other without affecting the grammaticality of the sentence. We generate mutated NLI pairs by finding occurrences of the prepositions in our list and randomly replacing them with other prepositions in the list. Our list consists of a set of locatives\footnote{Locative prepositions are those that denote place or position: e.g., \textit{in, on, near}, etc.} and several other manually-selected prepositions that are not strictly locatives but are likely to be substitutable (\textit{about}, \textit{for}, \textit{to}, \textit{with}, \textit{without}).

\paragraph{Comparatives} Comparatives express qualitative or quantitative differences between entities. For instance, a sentence that states \textit{A is more than B} and another that states \textit{B is more than A} lead to different inferences. We select a list of common comparatives (e.g., \textit{more/less, bigger/smaller}) and select pairs from MNLI that contain a comparative phrase in both the premise and the hypothesis. We mutate the sentences by negating the hypothesis.\footnote{We note that \citet{dasgupta2018evaluating} also focus on comparatives, but they exclusively look at artificial sentences containing \textit{more/less}.}

\paragraph{Quantification} The quantification task tests the understanding of natural language expressions of quantities, including common quantifiers (\textit{all, some}), number words (\textit{two, twenty}), and proportion (\textit{half, one-third, quarter}). We select NLI pairs that contain at least one quantifier in both the premise and the hypothesis, and again apply mutations by negating the hypothesis.

\paragraph{Spatial Expressions} The spatial expressions task probes the understanding of words that denote spatial relations between entities. Changing the spatial configuration often leads to different inferences; for instance, \textit{A is to the left of B} implies that \textit{B is to the right of A}, but not that \textit{A is to the right of B}. We select a set of words that describe spatial configurations which are not necessarily prepositions (e.g., \textit{left, right, close, far}). Again, we find MNLI pairs containing these words and negate to generate mutated pairs. 

\paragraph{Negation} This task probes whether models are able to understand negations, in particular explicit negation using the word \textit{not}, lexical negation using antonyms, and the interaction between them. We first identify premise-hypothesis pairs from the MNLI dataset that contain antonym pairs (e.g., \textit{dirty} appears in $p$ and \textit{clean} in $h$) and generate all possible patterns of negation with the two mutation strategies: swapping antonyms and adding explicit negation. That is, we use each of lexical negation, explicit negation, and their combination to mutate the premise and/or the hypothesis. We generate all 16 possible patterns of negation for a given premise-hypothesis $(p,h)$ pair. For each of $p$ and $h$ we can either apply or not apply each of four possible mutations: lexical negation, explicit negation, both, and none.

\subsection{Annotation}

We recruit human annotators on Amazon Mechanical Turk to produce the final labels for the heuristically-generated datasets described above. 
We collect three labels per sentence (or per pair of sentences for EOS and NLI probing sets). We use the majority label in our final dataset, and discard examples on which there is no majority consensus. For more details about our annotation protocol, including compensation, refer to Appendix~\ref{app:annotation}. 

\paragraph{Acceptability Tasks}
Human annotators are presented with a single (mutated or unmutated) sentence and are given the options \{\textit{natural}, \textit{unnatural}, \textit{neither}\}. We discard sentences in which the majority label does not agree with our expected label. That is, we only include mutated sentences with a majority label of \textit{unnatural} and unmutated sentences with a majority label of \textit{natural}. We collect around 500 annotated examples with balanced label ratio for each probing set. We release our sentences in small batches until we have approximately 250 \textit{unnatural} examples per task. To create the final dataset, we pool all answers from all batches and take a subset of the \textit{natural} sentences so that the label ratio is balanced, prioritizing examples with perfect inter-annotator agreement.

\paragraph{Natural Language Inference Tasks}
For the NLI tasks, we collect common-sense entailment judgments from annotators on a 5-point Likert scale on which 1 denotes `definitely contradiction' and 5 denotes `definitely entailment', following \citet{zhang2017ordinal}. This finer-grained scale is intended to avoid confounds arising from borderline cases. Except for the use of scaled judgments, our instructions follow the MNLI guidelines.
Specifically, our instructions said to assume that the sentences co-refer and that the first sentence ($p$) states a true fact, describes a scenario, or expresses an opinion, and to then indicate how likely it is that the second sentence ($h$) is also true, describes the same scenario, or expresses the same opinion.

Annotators could also select an option indicating that one or both of the sentences did not make sense; we discarded $(p,h)$ pairs for which at least one annotator chose this option. We map judgments of 5 and 4 to \textit{entailment}, 3 to \textit{neutral}, and 2 and 1 to \textit{contradiction}, and treat the majority label as the correct label after this mapping.

\paragraph{Agreement and Quality Control}
In constructing our final evaluation sets, we removed examples on which there was no majority consensus. For the binary acceptability tasks, we manually prefiltered sentences that were felicitous even after the heuristic modification. For the NLI tasks, we removed pairs that contained ungrammatical sentences that were not flagged by annotators via manual postfiltering. See Appendix~\ref{app:annotation} for more details. 

\section{Experimental Design}

\subsection{Pretraining Architecture}\label{sec:architecture}

Since our focus is on comparing differences in pretraining objectives, we fix the architecture for all sentence encoders. We use the pretrained character-level convolutional neural network (CNN) from ELMo \cite{peters2018deep} that replaces word embeddings (see \citet{friends} or \citet{tenney2019what} for similar usages of the CNN layer). This acts as a base input layer that uses no information beyond the word, and allows us to avoid potentially difficult issues surrounding unknown word handling in transfer learning. 

We feed the word representations to a 2-layer 1024d bidirectional LSTM \cite{hochreiter1997long}.
A downstream task-specific model sees both the top-layer hidden states of this model and, through a skip connection, the original representation of each word.
We train a version of this model on each task in Section~\ref{sec:pretraining-tasks}. Additional experimental details are in given in Appendix~\ref{app:exp-details}. Our codebase is open-source\footnote{\url{https://github.com/jsalt18-sentence-repl/jiant}} and built using AllenNLP \citep{Gardner2017AllenNLP} and PyTorch \citep{paszke2017automatic}.

\paragraph{Classification Tasks} For classification pretraining tasks (NLI, DisSent), we use an attention mechanism inspired by BiDAF \citep{Seo2016BidirectionalAF}. Given the sequence of output states of the core BiLSTM for both sentences in an example, we compute dot-product based attention between all pairs of words between the sentences to form a sequence of attention-contextualized word representations. We use an additional BiLSTM followed by max-pooling to obtain an attention-contextualized vector representation of each sentence $h_1$ and $h_2$. We use the \textit{heuristic matching} feature vector  $[h_1 ; h_2; h_1 \cdot h_2 ; | h_1 - h_2 | ]$  \citep{mou-EtAl:2016:P16-2} as input to an MLP. 

\paragraph{Sequence-to-Sequence Tasks} For sequence-to-sequence pretraining tasks (machine translation and skip-thought), we use a single-layer 1024d LSTM as the decoder, initialized with the max-pooled output of the encoder. We use a linear projection bottleneck layer to reduce the dimension of the output of the decoder by half before the output softmax layer.

\subsection{Pretraining Tasks}
\label{sec:pretraining-tasks}

Our main experiments compare seven pretraining tasks which we believe capture different aspects of linguistic meaning and which yield reasonable performance when used on a benchmark task such as MNLI.\footnote{Around 70\% development set accuracy; see Appendix~\ref{app:mnli-dev}.} For our purposes, a \textit{task} is a dataset-training objective pair. We attempt to select a set of tasks diverse enough to highlight performance differences due to pretraining objectives. We additionally report results using BERT \citep{devlin2018bert} (base, uncased) to demonstrate that our probing sets prove challenging even for state-of-the-art models.

\paragraph{Language Modeling} We train a left-to-right word-level language model on BWB, which was successfully used by \citet{peters2018deep} for pretraining sentence encoders. Because language modeling is trivial for a bidirectional LSTM, we follow \citet{peters2018deep} by training separate forward and backward two-layer 1024d language models and concatenate their hidden states as token representations.

\paragraph{Skip-Thought} 
Drawing from \citet{kiros2015skip} and \citet{W17-2625}, we train a sequence-to-sequence model on skip-thought, which is a task of generating the next sentence in the discourse given the previous sentence. We use the learned encoder as our sentence encoder. Since this objective requires running text, we use sentences from WikiText-103 as training data.

\paragraph{CCG Supertagging} We train a model to predict the Combinatory Categorial Grammar (CCG) supertag for each word, with sentences from CCGBank \cite{CCG}. Supertags are similar to part-of-speech tags but capture more syntactic context \citep[``almost-parsing'';][]{bangalore1999supertagging}.

\paragraph{Discourse (DisSent)} We train a model on DisSent \cite{jernite2017discourse, nie2017}, which is an unsupervised task of predicting the discourse marker (e.g., \textit{and}, \textit{because}, or \textit{so}) that connects two clauses. We train our model on a dataset created from WikiText-103 following \citet{nie2017}'s protocol, which involves extracting pairs of clauses with a specific dependency relation.

\paragraph{Natural Language Inference}
Inspired by \citet{Conneau2017SupervisedLO}, we use the MNLI dataset for NLI pretraining. The task is to predict the entailment label for premise-hypothesis pairs; the possible labels are \textit{entailment, contradiction, neutral}.
 
\paragraph{Machine Translation} 
We train a sequence-to-sequence machine translation model with attention on WMT14 English-German \citep{bojar2014findings} and take the encoder as our sentence encoder. \citet{mccann2017learned} previously showed that pretraining an encoder on translation led to good performance on downstream NLP tasks.

\paragraph{Image-Caption Matching} We train a model on the task of grounding sentences to the images they describe. We use image-caption pairs from the MSCOCO dataset \cite{lin2014microsoft} with an objective that minimizes the cosine distance between sentence representations and corresponding image features, as described in \citet{DBLP:journals/corr/KielaCJN17}.

\subsection{Classifiers for Probing Tasks}
\label{sec:task-specific-classifiers}

To probe the sentence encoders pretrained on the different objectives, we freeze the weights of the encoder after pretraining and train an additional model using the outputs of the fixed encoder as inputs. We describe the implementation details for the NLI and acceptability probing sets below.

\paragraph{NLI Tasks} For NLI-type probing, we train an NLI model on top of the representations produced by the pretrained sentence encoder that uses an attention mechanism inspired by \citet{Seo2016BidirectionalAF} that computes attention between all pairs of words in the two sentences (described in more detail in Section~\ref{sec:architecture}). We train this component on MNLI and evaluate directly on our NLI probing datasets with no further dataset-specific training. 

\paragraph{Acceptability Classification Tasks} 
For all acceptability tasks except the EOS task, we take the sequence of hidden state outputs from the pretrained encoder as the sentence representation. We aggregate this sequence into a single vector via max-pooling and train a 512d MLP on top of the resulting vector.
For the EOS task, we also use max-pooling on each sentence in the pair. We then concatenate the resulting vectors and train an MLP on top of the joint representation.\footnote{We tried training a general acceptability model using CoLA and evaluating directly on our acceptability tasks, as an analogous evaluation setup to the NLI tasks, but all models performed around chance under this setup. This is likely due to the intrinsic difficulty of CoLA for our base model, as suggested by low performance from similar models (``GLUE Baselines'') on \url{https://gluebenchmark.com}.} 
Each task has around 400 training examples (see Appendix~\ref{app:annotation}). Due to their small size, we use 10-fold cross validation where each fold is used as the test set exactly once, and report the average test set accuracy.

\paragraph{BERT}

For NLI-type probing tasks, we use the fine-tuned MNLI classifier from \cite{devlin2018bert}\footnote{\url{https://github.com/google-research/bert}}. For the acceptability classification tasks, we fine-tune the model by adding a sequence-level classifier on top of the pretrained BERT model. The sequence-level classifier is a linear layer that takes in as input the final hidden vector corresponding to the first input token as aggregate representation in the input sequence, and then classifies to the required number of classes for the task, where the label probabilities are computed with a standard softmax. The BERT fine-tuning setup allows a classification output to be indicated with a CLS token. Pairs of sequences are indicated with a SEP token between the pairs. All parameters are fine-tuned jointly to maximize the log-probability of the correct label while the hyperparameters are the same as in pretraining.

\begin{figure*}[ht!]
\centering
	\includegraphics[width=\linewidth]{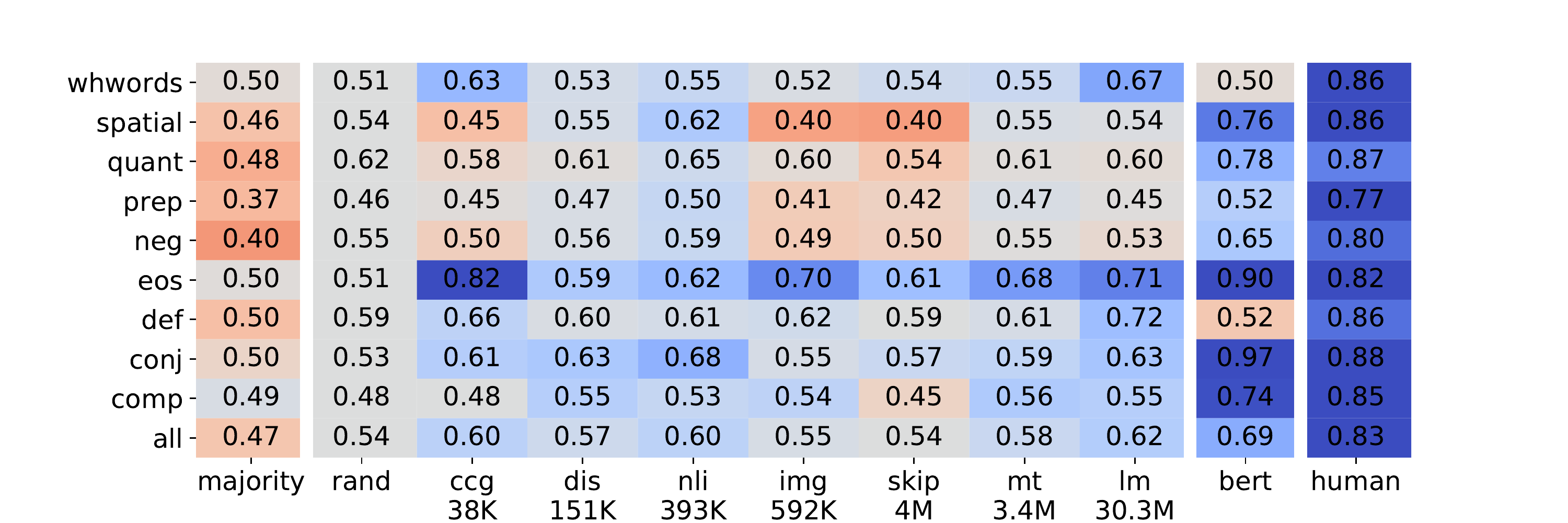}
	\caption{Accuracy for each pretraining task on each probing set. The leftmost column shows the majority-class baseline, and the rightmost column shows individual annotator accuracy on the final probing set. Blue denotes performance improvement over randomly initialized encoder baseline and orange denotes performance decrease.}
\label{table:big-table}
\end{figure*}

\subsection{Variation from Random Restarts}

In order to calibrate the degree of variation that can be expected due to random restarts, we run each of our probing tasks on five different random initializations of the sentence encoder weights. These sentence encoders were not pretrained, and we trained MLPs for each probing task on top of the randomly initialized sentence encoders. The expectation is that if pretraining has measurable effects on the probing results, the variance across different pretrained models would be greater than the variance across random restart models.
Across five random restarts, the average standard deviation across our probing set was around 1 percentage point. The mean and standard deviation for each probing task are reported in Appendix~\ref{app:random}.

\section{Results}
\subsection{Overall Performance}
Figure \ref{table:big-table} shows the performances of models trained on each pretraining task on our probing datasets. We also provide comparison with a randomly initialized encoder with no pretraining, which is known to be a strong baseline \citep{friends}. We observe that different pretraining tasks have different strengths and weaknesses; there is no single pretraining task that achieves the best (or worst) performance across the board. This implies that even the best encoders, such as BERT, are unable to capture function word semantics fully, and suggests further research into combining advantages of different tasks. Furthermore, most models are far from human performance, with only a few exceptions (e.g., BERT on conjunctions). This demonstrates that our probing datasets serve as useful challenge sets, in addition to permitting fine-grained analysis. 

Looking into each probing set in more detail, we see several intuitive patterns on how pretraining might affect probing performance. Among the pretrained models (not including BERT), the NLI model outperformed all other models on the negation\footnote{We additionally find that this improvement is specifically due to the NLI model's capacity to understand explicit negation using \textit{not}, rather than lexical negation with antonymy. See Appendix~\ref{app:neg-subset} for differences between negation subtypes.} and conjunction tasks, both of which involve words that play central roles in inferential reasoning. The CCG model yields the best result for EOS, which could be attributed to the task's emphasis on structure; it is the only task where the target labels directly represent compositional structure.

Surprisingly, we find that pretraining can sometimes hurt performance. For instance, pretraining on the skip-thought objective or the image captioning objective appears to hurt performance on spatial relations and prepositions (compared to the performance of a randomly initialized encoder). In fact, for many probing sets, the choice of pretraining task affects whether it helps or hurts performance; for instance, pretraining on NLI helps with negation, whereas pretraining on image-caption matching and CCG lowers performance. This suggests that pretraining can be helpful, but \textit{only} helpful if we pretrain on a task that provides useful information in solving the target task. For instance, in Section~\ref{sec:analysis} we discuss how the image-caption matching objective may bias models to discard information about certain preposition senses. Overall, we observe that language modeling is the most useful pretraining task, although CCG and NLI perform comparably (within random variation range). The strong performance of language modeling aligns with its effectiveness for pretraining models that achieve state-of-the-art NLP results. However, we note that CCG, our most syntactic task, achieves competitive results with the smallest number of training examples (an order of magnitude less data than NLI and two orders less than language modeling), suggesting that incorporating syntactic knowledge might yield more efficient learning of function word semantics.

We furthermore see that many of our probing sets are challenging even for BERT---although BERT substantially improves performance on many probing sets, and obtains superhuman performance on conjunctions and EOS,\footnote{We speculate that this might be an effect of the next-sentence classification task that BERT is pretrained on.} it also shows clear weaknesses in several probing sets (e.g., \textit{wh}-words and definite/indefinite articles) where it is outperformed even by a randomly initialized baseline with no pretraining.

\begin{figure*}[ht!]
\centering
	\includegraphics[width=1\linewidth]{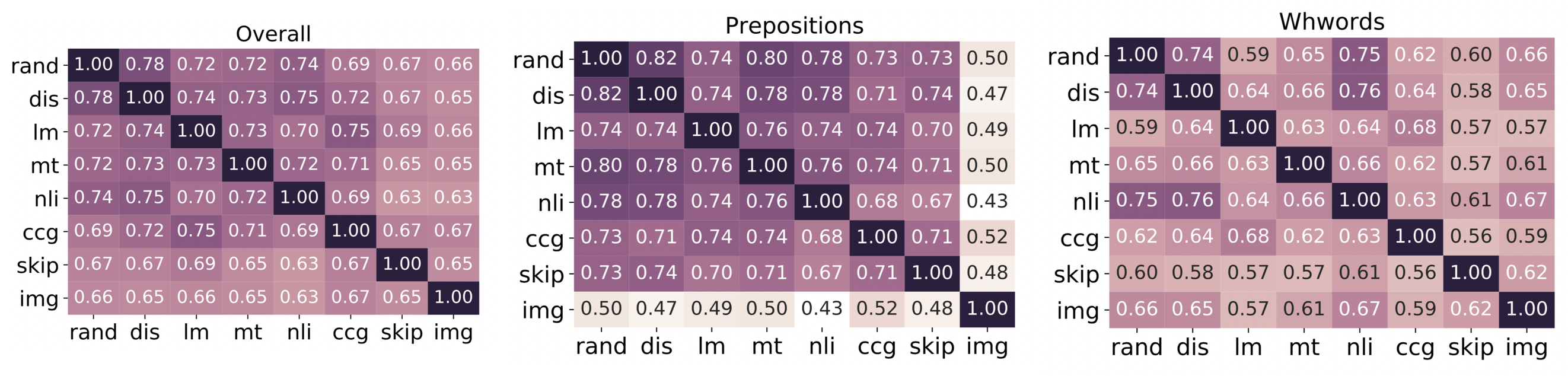}
	\caption{Prediction overlap on the probing tasks for models trained on different pretraining tasks (i.e., how often models make identical predictions on a particular probing set).}
	\label{fig:correlations}
\end{figure*}

\subsection{Correlations between Pretaining Tasks}
To further investigate whether our probing sets differentiate between pretraining objectives, we look into correlations between the model predictions; given two pretraining tasks $i$ and $j$, how often does a model trained on $i$ make the exact same prediction as a model trained on $j$? Figure~\ref{fig:correlations} shows the correlations across all probing sets in aggregate, and for the \textit{wh}-words and prepositions sets specifically (see Appendix~\ref{app:overlap} for all sets). 

We observe that models pretrained on different tasks do make different predictions overall, with image-caption matching and skip-thought being the tasks that make predictions that deviate the most from others (left). NLI and image-caption matching are the least correlated pair of tasks among all. The difference between image-captioning and other tasks is the most prominent in the preposition probing set; it makes predictions that are only weakly correlated with others (middle). We hypothesize that this is due to the duality of preposition semantics; most prepositions have both concrete and abstract senses, and the image model is biased to focus on the former. 

To illustrate, consider the preposition \textit{below}, which can denote a spatial configuration (e.g., \textit{the boots end \textbf{below} the knee}) or an abstract relation (numeric or qualitative comparison; e.g., \textit{her score is \textbf{below} sixty}). In the preposition dataset, \textit{below} occurs 17 times, 11 of which are spatial and 6 abstract. For the spatial usage, both MNLI and image-caption models have 64\% accuracy (7/11). The NLI model shows 50\% accuracy for pairs containing abstract uses (3/6), but the image-captioning model answers none of them correctly (0/6). Here is an example of a numeric usage of \textit{below} that the NLI model answered correctly but the image model answered incorrectly:

P: \textit{Only those whose incomes do not exceed 125 percent of the federal poverty level qualify $\dots$}

H: \textit{Those whose incomes are \textbf{below} 125 percent qualify $\dots$} \hfill (P$\to$H)

\noindent The image model's bias towards the spatial usage is intuitive, since the numeric usage of \textit{below} (i.e., as a counterpart to \textit{exceed}) is difficult to learn from visual clues only. This concrete-abstract duality, which is not specific to \textit{below} but common to most other prepositions \citep{schneider2018preps}, may partially explain why the image-caption model behaves so differently from all other models, which are not trained on a multimodal objective.

\subsection{Data Size and Genre Effects}
\label{sec:analysis}
As can be seen from the varying sizes of the pretraining dataset reported in Figure~\ref{table:big-table}, seeing more data during pretraining does not imply better performance on probing tasks. Also, as noted before, the fact that pretraining can hurt performance suggests that if the task is not the ``right'' task, adding more datapoints during pretraining can lead models to learn counterproductive representations. 

Another potential confound is vocabulary overlap between pretraining and probing task datasets. Since all pretraining task datasets have different sets of vocabulary, the variance in the results could be attributed to the number of words in the probing set that were already seen in pretraining. To investigate this possibility, we compute the ratio of overlapping words between the pretraining and probing datasets. A regression analysis shows that vocabulary overlap overall is not a significant predictor of performance on the probing set ($p = 0.39$). No single probing set performance was significantly affected by vocabulary overlap either (all $p>.05$ after Bonferroni correction for multiple comparisons).

\section{Related Work}
An active line of work focuses on ``probing'' neural representations of language. \citet{ettinger2016probing,ettinger2017towards,zhu2018exploring}, \textit{i.a.,} use a task-based approach similar to ours, where tasks that require a specific subset of linguistic knowledge are used to perform qualitative evaluation. \citet{gulordava2018colorless}, \citet{giulianelli2018under}, \citet{W18-5409}, and \citet{jumelet2018language} make a focused contribution towards a particular linguistic phenomenon (agreement, ellipsis, negative polarity). Using recast NLI, \citet{N18-2082} probe for semantic phenomena in neural machine translation encoders.  \citet{W17-5410,W17-5405,ribeiro2018semantically} use similar strategies to our structural mutation method, although their primary goal was to break existing systems by adversarial modifications rather than to compare different models. \citet{ribeiro2018semantically} and our work both test for proper comprehension of the modified expressions, but our modifications are designed to induce semantic changes whereas their modifications are intended to preserve the original meaning. Our strategy is close to that of \citet{naik2018stress}, but our modifications are more constrained and lexically targeted. 

The design of our NLI-style probing tasks follows the recent line of work which advocates for NLI as a general-purpose format for diagnostic tasks \cite{white2017inference,poliak2018collecting}. This idea is similar in spirit to \citet{McCann2018decaNLP}, which advocates for question answering as a general-purpose format, to edge~probing~\cite{tenney2019what} which probes for syntactic and semantic structures via a common labeling format, and to GLUE~\cite{wang2018glue} which aggregates a variety of tasks that share a common sentence-classification format. The primary difference in our work is that we focus specifically on the understanding of function words in context. We also present a suite of several tasks, but each one focuses on a particular structure, whereas tasks proposed in the works above generally aggregate multiple phenomena. Each of our tasks isolates each function word type and employ a targeted modification strategy that gives us a more narrowly-focused, informative scope of analysis.

\section{Conclusion}

We propose a new challenge set of nine tasks that focus on probing function word comprehension. Although we use our challenge set to compare the effects of pretraining, the probing sets themselves are architecture- and evaluation setup-agnostic. The results show that models pretrained with different objectives do generate different predictions (e.g., image models have a bias towards concrete preposition senses), and that no single objective leads to models that perform best or worst across all tasks. This suggests that there are `gaps' in the linguistic knowledge learned from a single pretraining objective that could be complemented by others, and calls for further research into how different pretraining objectives could be productively combined. In addition to contributing to the discussion of finding effective pretraining tasks, we hope that our exploratory study initiates further discussions about modeling function words and their contribution to compositional meaning. 

\section*{Acknowledgments}
The work was conducted at the 2018 Frederick Jelinek Memorial Summer Workshop on Speech and Language Technologies, and supported by Johns Hopkins University with unrestricted gifts from Amazon, Facebook, Google, Microsoft and Mitsubishi Electric Research Laboratories.

\bibliography{naaclhlt2019} 
\bibliographystyle{acl_natbib}

\section*{Appendix}

\appendix
\section{Experimental Details}
\label{app:exp-details}

\paragraph{Image-Caption Matching} We train on image-caption pairs from the MSCOCO dataset \citep{lin2014microsoft} to minimize the cosine distance between representations of the sentence and corresponding image. Specifically, we encode the sentence with the BiLSTM encoder. We use a Resnet-101 \citep{he2016deep}, a CNN pretrained on ImageNet to obtain a 1024-dimensional feature representation for the image. We linearly transform the encoder output of the sentence to the size of the image representation and use a cosine embedding loss against the two vectors, i.e., minimize the cosine distance between two representations to allow mapping sentences to their corresponding images.
 
\paragraph{Regularization} 
We regularize with dropout with $p=0.2$. Dropout is placed after the input layer, each LSTM layer, and each MLP layer in the task-specific classifier.

\paragraph{Preprocessing}

We use Moses tokenizer with a maximum sequence length of 40 tokens. Because we used a character-based word encoder, we have no word-level vocabulary, 
except for sequence-to-sequence tasks, where we use an output vocabulary of 20,000 tokens. For translation, we use BPE tokenization; for skip-thought we use the Moses tokenizer.

\paragraph{Training Details}

We optimize using AMSGrad \citep{j.2018on} with a learning rate of 1e-3 for text generation tasks and 1e-4 otherwise.
We evaluate on the validation set every 1,000 iterations and stop training if we fail to get a best result after 20 evaluations.
We multiply the learning rate by 0.5 whenever validation performance fails to improve for more than 4 validation checks. We also stop training if the learning rate falls below 1e-6.
At the end of training, we load the best checkpoint.

\paragraph{Acceptability task evaluation}
For the acceptability tasks, we use a 10-fold cross-validation evaluation setup because we are training task-specific classifiers for each probing task and the datasets are small. We split each dataset into 10 folds with balanced label ratio within each fold, and test on each fold using the other 9 as train and development sets (8 folds train, 1 fold dev). The accuracy reported in the paper for the acceptability tasks is test accuracy averaged across all folds.

\section{MNLI Development Set Accuracy for Pretrained Models}
\label{app:mnli-dev}

\begin{table}[h]
    \centering
    \begin{tabular}{l|l}
                & MNLI (dev) \\ \toprule
    Random    & 73.8  \\
    CCG    & 69.6\\
    DisSent  & 73.6  \\
    MNLI & 75.6 \\
    IMG & 62.6 \\
    Skip & 67.4  \\
    MT & 72.0 \\
    LM & 72.6 \\
    \end{tabular}
    \caption{MNLI development set accuracy for each pretrained model.}
    \label{table:mnli-dev}
\end{table}

\section{Annotation Protocol}
\label{app:annotation}
We recruited three annotators per sentence or sentence pair on Amazon Mechanical Turk to control the quality of the labels for our heuristically generated datasets. For the acceptability judgment task sentences, individual sentence or sentence pair example was presented to the annotators and they were asked to choose between the options \textit{natural, unnatural, neither}, after reading the given example. The examples were presented in sets of five sentences (individual sentence tasks) or three sentence pairs (sentence pair tasks) in random order, with the option to stop at any point during the process. The annotators were compensated with $\$.1$ per five sentences (or three sentence pairs). For NLI task sentences, the annotators were presented with six sentence pairs, for which they were asked to provide judgment on a five-point scale about the inferrability of the second sentence from the first. The annotators were compensated with $\$.1$ per six sentence pairs. See Table~\ref{table:agreement-stats} for inter-annotator agreement and the final size of the dataset.

\begin{table}[h]
    \centering \small
    \begin{tabular}{l|ccc|c}
                    & Agree & Unan.  & Accuracy & Size \\ 
\toprule
    Negation    &   60.2 &  40.3  & 80.1 & 591\\
    Spatial     &   71.1 &   56.7 & 85.6 & 157 \\
    Quant.      &   73.8  & 60.7  & 86.9 & 323 \\
    Comp.       &   70.0  & 55.1  & 85.0 & 89 \\
    Prep        &   54.7 &  32.1  & 77.4 & 358 \\
    \midrule
    \textit{wh}-words & 72.5 & 58.7 & 86.2 & 584 \\ 
    Def.  & 72.0 & 58.1 & 86.0 &508 \\
    Coord. & 75.3 & 62.9 & 87.6 & 456 \\ 
    EOS    &  64.9 & 47.3 & 82.4 & 478 \\ 
    \end{tabular}
    \caption{Pairwise inter-annotator agreement ($n=3$), \% of examples with unanimous agreement, and individual annotator accuracy according to the expected label for each task in the final probing dataset.}
    \label{table:agreement-stats}
\end{table}

\section{List of Prepositions Used for the NLI Probing Set}
\label{app:preplist}
\textit{about, above, across, after, against, ahead of, all over, along, among, around, at, before, behind, below, beneath, beside, by, for, from, in, in front of, inside, inside of, into, near, nearby, next to, on, on top of, out of, outside, outside of, over, past, through, to, under, up, within, with, without}

\section{Random Initialization Variance}
\label{app:random}
\begin{table}[h]
    \centering
    \begin{tabular}{l|l|}
                & $\mu$ $(\pm \sigma)$ \\ \toprule
    Prep    & $46.14$ $(\pm 0.89)$\\
    Negation    &  $53.78$ $(\pm 0.84)$\\
    Spatial    & $56.18$ $(\pm 1.68)$ \\
    
    Quant.    & $61.1$ $(\pm 1.55)$ \\
    Comp.    & $51.22$ $(\pm 2.31)$\\\midrule 
    \textit{wh}-words & $51.37$ $(\pm 0.36)$\\ 
    def/indef articles & $57.77$ $(\pm 0.97)$\\
    coord. & $54.39$ $(\pm 0.96)$\\
    EOS    & $52.74$ $(\pm 1.30)$\\ 
    \end{tabular}
    \caption{Mean and standard deviation of probing task performance across five different random initializations.}
    \label{table:random-restarts}
\end{table}

\begin{figure*}[t!]
\centering
	\includegraphics[width=1\linewidth]{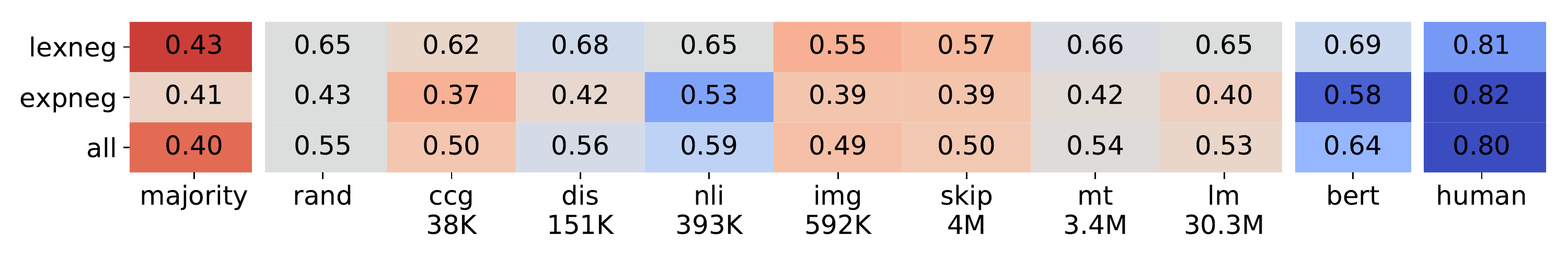}
	\caption{Accuracy of each pretrained model on subsets of the negation probing set. \textit{neg} is the accuracy for the whole negation probing set. \textit{lexneg} shows accuracy for the subset of sentence pairs negated using antonyms and \textit{expneg} sentences explicitly negated using \textit{not}. The leftmost column shows the majority-class baseline, and the rightmost column shows individual annotator accuracy on the final evaluation set. Blue denotes performance improvement over randomly initialized encoder baseline and orange denotes performance decrease.}
\label{table:neg-table}
\end{figure*}

\section{Subset Accuracy for the Negation Probing Set}
\label{app:neg-subset}
In Figure~\ref{table:neg-table}, we see that the NLI model's improvement on the negation probing set mostly derives from its improvement on explicit negation rather than lexical negation.

\section{Prediction Overlap between Models}
\label{app:overlap}
\begin{figure*}[h]
\centering
	\includegraphics[width=1\linewidth]{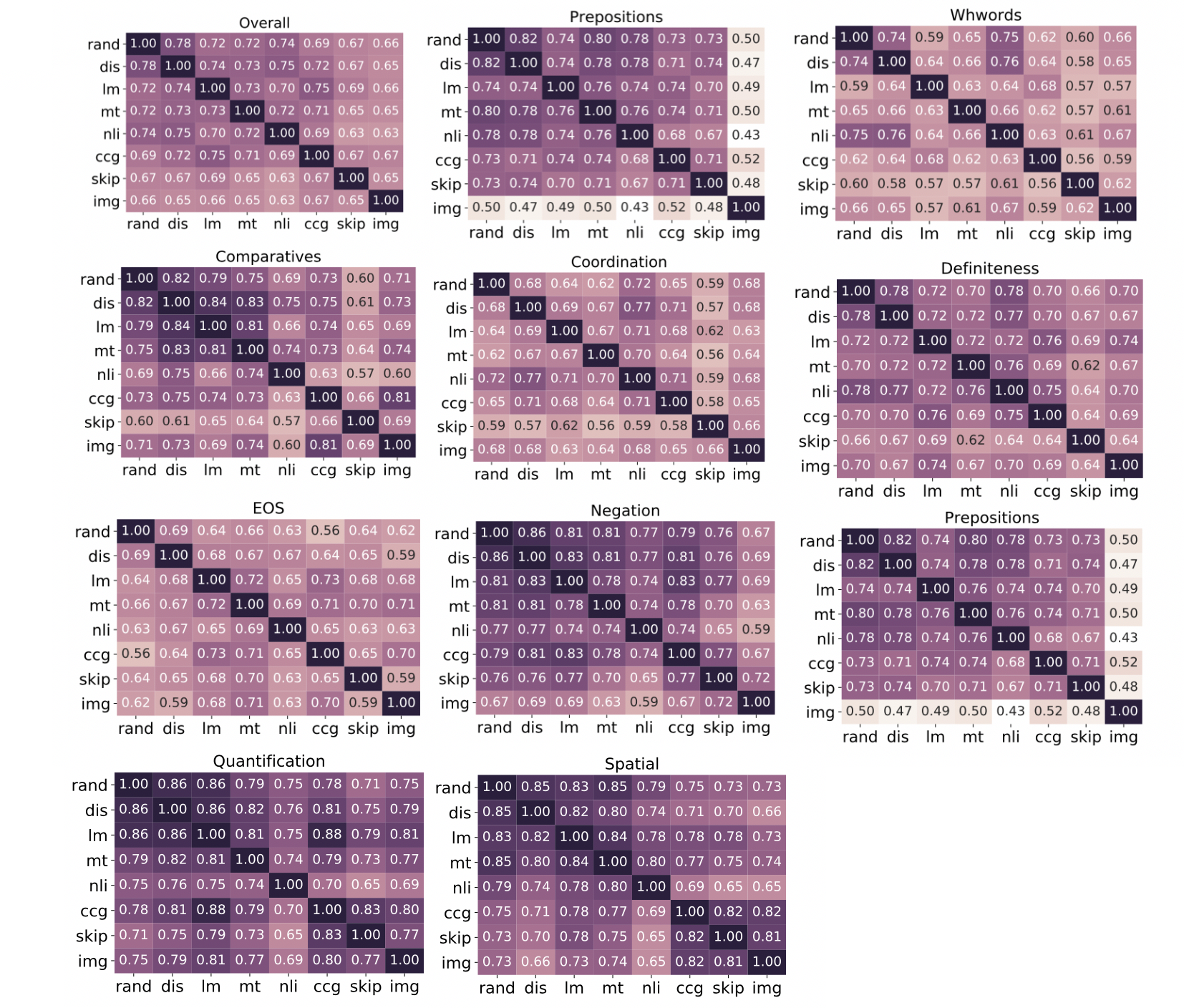}
	\caption{Heatmaps of prediction overlap for all probing tasks, between models pretrained with different objectives.}
\label{fig:heatmaps}
\end{figure*}

We show prediction overlap heatmaps for all probing tasks in Figure~\ref{fig:heatmaps}.

\end{document}